\documentclass[wcp]{jmlr}

\usepackage{longtable}

\usepackage{booktabs}
\usepackage{graphbox, xcolor}
\usepackage{hyperref}
\usepackage{booktabs} 
\usepackage{amsmath}
\usepackage{amssymb}
\usepackage{mathbbol}
\usepackage{arydshln}
\usepackage{algorithm}
\usepackage{algorithmic}
\newcommand{\argmax}{\mathop{\rm arg~max}\limits}
\newcommand{\argmin}{\mathop{\rm arg~min}\limits}

\usepackage{lineno}
\makeatletter
\let\Ginclude@graphics\@org@Ginclude@graphics 
\makeatother

\title[Generative Semi-supervised Learning with Meta-Optimized Synthetic Samples]{Generative Semi-supervised Learning \\with Meta-Optimized Synthetic Samples}

 \author{\Name{Shin'ya Yamaguchi} \Email{shinya.yamaguchi@ntt.com}\\
  \addr NTT / Kyoto University}

\begin{document}

\maketitle

\begin{abstract}
Semi-supervised learning (SSL) is a promising approach for training deep classification models using labeled and unlabeled datasets.
However, existing SSL methods rely on a large unlabeled dataset, which may not always be available in many real-world applications due to legal constraints (e.g., GDPR).
In this paper, we investigate the research question: \textit{Can we train SSL models without real unlabeled datasets?}
Instead of using real unlabeled datasets, we propose an SSL method using synthetic datasets generated from generative foundation models trained on datasets containing millions of samples in diverse domains (e.g., ImageNet).
Our main concepts are identifying synthetic samples that emulate unlabeled samples from generative foundation models and training classifiers using these synthetic samples.
To achieve this, our method is formulated as an alternating optimization problem: (i) meta-learning of generative foundation models and (ii) SSL of classifiers using real labeled and synthetic unlabeled samples.
For (i), we propose a meta-learning objective that optimizes latent variables to generate samples that resemble real labeled samples and minimize the validation loss.
For (ii), we propose a simple unsupervised loss function that regularizes the feature extractors of classifiers to maximize the performance improvement obtained from synthetic samples.
We confirm that our method outperforms baselines using generative foundation models on SSL.
We also demonstrate that our methods outperform SSL using real unlabeled datasets in scenarios with extremely small amounts of labeled datasets. 
This suggests that synthetic samples have the potential to provide improvement gains more efficiently than real unlabeled data.
\end{abstract}

\section{Introduction}
Semi-supervised learning (SSL) is a promising approach for training deep neural network models with a limited amount of labeled data and a large amount of unlabeled data.
Recent studies on SSL have shown that the labeling cost to achieve high-performance models can be significantly reduced by using the unlabeled dataset to train the models with pseudo-labeling and consistency regularization~\citep{bachman_NIPS14_learning_with_pseudo_ensembles,xie_NIPS20_UDA,sohn_NIPS20_fixmatch}.
For example, \cite{Wang_ICLR23_freematch} have reported that their SSL method can achieve 94.22\% test accuracy on CIFAR-10 with only one label per class.
This indicates that modern SSL methods can realize practical models with minimal labeling costs.
However, whether labeled or not, large-scale datasets are becoming more challenging to obtain and use for machine learning models due to privacy regulations (e.g., GDPR in the EU).

To train deep models in a situation where it is challenging to obtain datasets, recent studies using synthetic datasets from deep generative models have attracted much attention in the context of proxying real datasets~\citep{He_ICLR23_synthetic,van_ICML23_synthetic}.
This approach has been intensively discussed in the community\footnote{NeurIPS Synthetic Data Workshop (https://www.syntheticdata4ml.vanderschaar-lab.com/)} and regarded as a promising method for privacy protection since generative models such as GANs~\citep{Goodfellow_NIPS14_GANs} can produce realistic samples while guaranteeing certain differential privacy~\citep{Lin_ICML21_privacy_properties_of_gans}.
If the synthetic samples can be used as unlabeled datasets in SSL, we can train a high-performance model without real unlabeled datasets and privacy risks.
Thus, we investigate a research question: \textit{Can we train SSL models with synthetic unlabeled datasets instead of real ones?}\looseness-1

\begin{figure}[t]
    \centering
    \includegraphics[width=0.8\columnwidth]{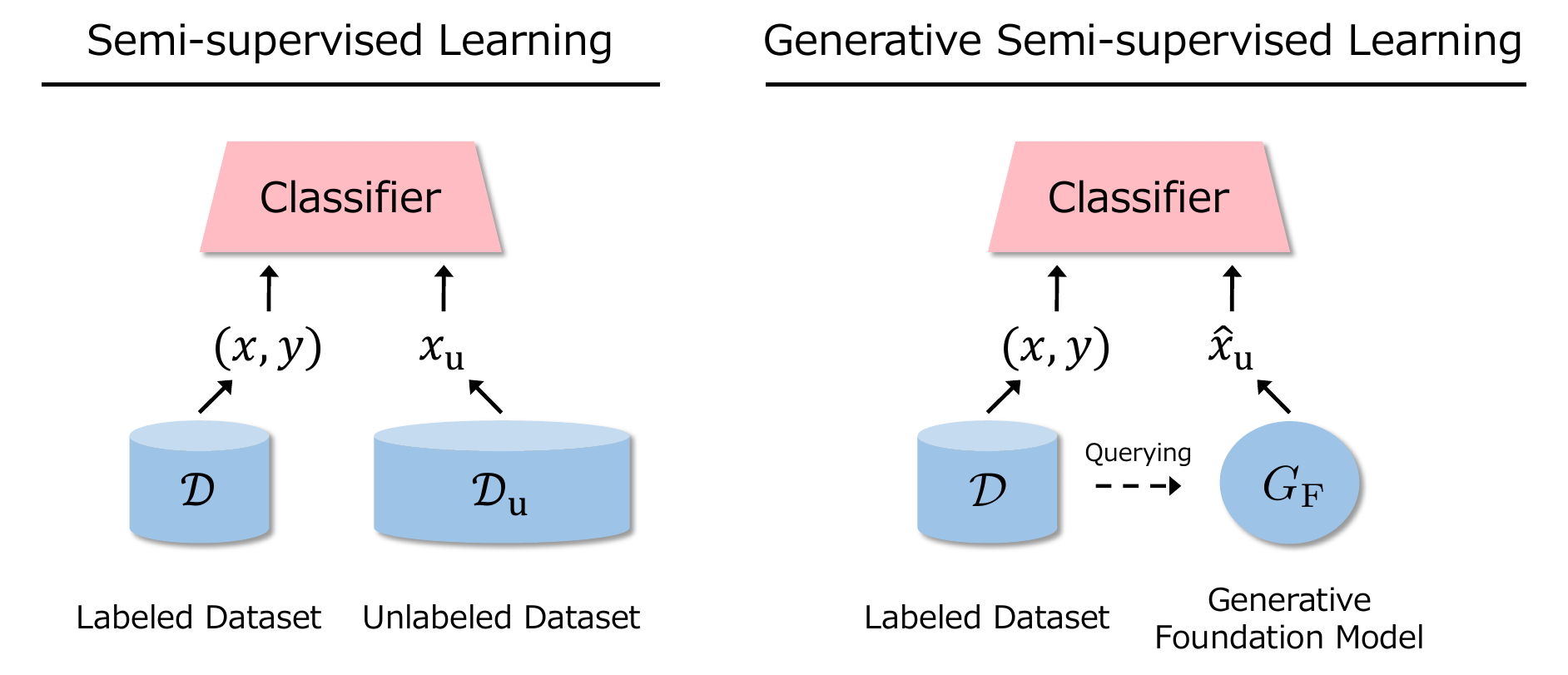}
    \caption{
    Comparison of semi-supervised learning (SSL) and generative semi-supervised learning (gSSL). In gSSL, we use a generative foundation model \(G_\mathrm{F}\) to compute unsupervised losses instead of a real unlabeled dataset \(\mathcal{D}_\mathrm{u}\). To this end, we generate synthetic unlabeled samples by querying \(G_\mathrm{F}\) with information of the labeled dataset \(\mathcal{D}\).
    }
    \label{fig:top}
\end{figure}

In this paper, we explore a new problem setting called \textit{generative semi-supervised learning} (gSSL), where the semi-supervised learners use synthetic unlabeled samples generated from a \textit{generative foundation model} instead of real unlabeled samples (Figure~\ref{fig:top}).
Generative foundation models are conditional generative models pre-trained on large external datasets containing millions of samples from diverse domains (e.g., ImageNet).
Thanks to recent advances~\citep{brock2018biggan,Sauer_SIGGRAPH22_styleganxl}, generative foundation models can accurately output synthetic samples in various domains from inputs of latent variables and conditional labels.
Therefore, we can expect the synthetic samples to perform as the unlabeled datasets in SSL when the training data space overlaps the data space estimated by the generative foundation models.

In gSSL, there are important challenges according to the following two concrete research questions: (i) \textit{How do we find optimal synthetic samples from the generative foundation models for SSL?} and (ii) \textit{How do we train models with synthetic samples that do not belong to the training class categories?}
For (i), since generative foundation models do not necessarily have the same class categories in the training datasets, we need to find synthetic samples from the generative models related to training datasets.
Furthermore, it is essential to find synthetic samples that can improve the classifier performance through the unsupervised loss of SSL.
For (ii), even if we can find helpful synthetic samples from the generative foundation models, it is not obvious how these samples can be optimally used to maximize the performance of the classifiers.
This is because the synthetic samples are matched to training datasets with respect to the domain (data space), not the class label spaces.
Since existing SSL methods assume that unlabeled samples belong to the training class categories, the mismatch between real and synthetic samples in the class label spaces can be detrimental to SSL models.

To address these two challenges, we propose a method called \textit{meta-pseudo semi-supervised learning} (MP-SSL).
MP-SSL consists of two techniques corresponding to the two research questions: (i) \textit{latent meta-optimization} (LMO) and (ii) \textit{synthetic consistency regularization} (SCR).
In LMO, we optimize latent variables that are input to generative foundation models to find synthetic samples that resemble unlabeled training data.
To find optimal synthetic samples for SSL models, LMO meta-optimizes the parameters to minimize the validation losses of the target classifier.
Furthermore, LMO also minimizes the gaps in the feature spaces between real and synthetic samples to align the domain gap and make the synthetic samples perform as unlabeled data.
SRC is a novel unsupervised loss term without the use of pseudo training labels.
Unlike existing SSL methods depending on real unlabeled data and the pseudo training labels, the SCR loss is designed as a feature regularization term.
This design choice is to avoid the negative effects of the synthetic samples caused by the mismatch of the class label spaces.
Specifically, SCR penalizes the feature extractors by maximizing the similarity between variations of a synthetic sample, which is inspired by consistency regularization~\citep{bachman_NIPS14_learning_with_pseudo_ensembles,xie_NIPS20_UDA,sohn_NIPS20_fixmatch}.
Since SRC is independent of the relationship between training and foundation label spaces, it can leverage the valuable information contained in synthetic unlabeled data to train the model without negative effects.
The training objective of MP-SSL is formalized as an alternating optimization problem of updating latent variables and updating training models through SSL with SCR.

To evaluate the effectiveness of MP-SSL, we conduct the experiments on multiple datasets by comparing MP-SSL with competitors, including P-SSL~\citep{Yamaguchi_arXiv22_PSSL}.
We also compare MP-SSL with SSL methods using real unlabeled datasets.
The results show that MP-SSL outperforms the real SSL methods when the labeled datasets are small.
This suggests that synthetic samples can promote more effective learning than real unlabeled samples, especially in cases where the number of labels is extremely small.
We believe this work will be the baseline for developing a new research area, generative semi-supervised learning.\looseness-1

Our contributions are summarized as follows.
\begin{itemize}
    \item We propose a new problem setting of SSL called generative semi-supervised learning (gSSL), where the unlabeled samples are provided by generative foundation models instead of real unlabeled datasets.
    \item We introduce a training method for gSSL called MP-SSL, which finds optimal synthetic samples performing as unlabeled data through meta-optimizing latent variables and trains a classifier with a feature regularization with the synthetic samples.
    \item We confirm that MP-SSL can outperform simple baselines of the gSSL setting and outperform SSL methods with real unlabeled datasets in small amounts of labels.
\end{itemize}

\section{Preliminary}\label{sec:preliminary}
\subsection{Problem Setting}\label{sec:problem_setting}
We consider a classification problem in which we train a neural network model \(f_{\theta} :\mathcal{X} \to \mathcal{Y}\) on a labeled dataset \(\mathcal{D}=\{(x^i,y^i) \in \mathcal{X}\times\mathcal{Y}\}^{N}_{i=1}\), where \(\mathcal{X}\) and \(\mathcal{Y}\) are the input and output label spaces, respectively.
In this setting, we can use a generative foundation model \(G_\mathrm{F}:\mathcal{Z}_\mathrm{F}\times\mathcal{Y}_\mathrm{F} \to \mathcal{X}_\mathrm{F}\), where \(\mathcal{Z}_\mathrm{F}\) is the latent space, \(\mathcal{Y}_\mathrm{F}\) is the foundation label space, and \(\mathcal{X}_\mathrm{F}\) is the output sample space.
We assume that \(G_\mathrm{F}\) is pre-trained on a large-scale dataset (e.g., ImageNet) and the output sample space \(\mathcal{X}_\mathrm{F}\) contains a subset \(\mathcal{X}^\prime\) related to \(\mathcal{X}\), i.e., \(\mathcal{X}_\mathrm{F} \supset \mathcal{X}^\prime \approx \mathcal{X}\).
An input latent variable \(z\in \mathcal{Z}_\mathrm{F}\) is sampled from a standard Gaussian distribution \(\mathcal{N}(0,I)\).
\(f_{\theta}\) is defined by a composition of a feature extractor \(g_{\psi}\) and a classifier \(h_{\omega}\), i.e., \(f_{\theta} = h_\omega \circ g_\psi\) and \(\theta = [\psi,\omega]\).
To validate \(f_\theta\), we can use a small validation dataset \(\mathcal{D}_\text{val}=\{(x^i_\text{val}, y^i_\text{val}) \in \mathcal{X}\times\mathcal{Y}\}^{N_\text{val}}_{i=1}\), which has no intersection with \(\mathcal{D}\) (i.e., \(\mathcal{D}\cap\mathcal{D}_\text{val}=\emptyset\)).

\subsection{Semi-supervised Learning}\label{sec:semi-supervised-learning}
Given a labeled dataset \(\mathcal{D}\) and an unlabeled dataset \(\mathcal{D}_\mathrm{u}=\{x^i\in\mathcal{X}\}^{N_\mathrm{u}}_{i=1}\), SSL to train \(f_\theta\) is formulated as the following minimization problem.
\begin{eqnarray}\label{eq:ssl_loss}
    &\underset{\theta}{\min}~\mathcal{L}(\theta)+\lambda_\mathrm{u} \mathcal{L}_{\mathrm{u}}(\theta),\\
    &\mathcal{L}(\theta) = \frac{1}{N} \sum_{(x, y) \in \mathcal{D}} \ell(f_\theta(x), y)\\
    &\mathcal{L}_\mathrm{u}(\theta) = \frac{1}{N_\mathrm{u}} \sum_{x_\mathrm{u} \in \mathcal{D}_\mathrm{u}} \ell_\mathrm{u}(f_\theta(x_\mathrm{u}))
\end{eqnarray}
where \(\ell\) is a supervised loss for a labeled sample (e.g., cross-entropy loss), \(\ell_\mathrm{u}\) is an unsupervised loss for an unlabeled sample \(x_\mathrm{u}\), and \(\lambda_\mathrm{u}\) is a hyperparameter for balancing \(\mathcal{L}\) and \(\mathcal{L}_\mathrm{u}\).
SSL assumes a large amount of unlabeled data (i.e., \(N\ll N_\mathrm{u}\)).
This assumption has long been justified on the premise that the difficulty of the dataset creation is centered on labeling, and the collection of unlabeled data can be easily done~\citep{chapelle_MIT06_ssl_survey}.
However, unlabeled data are often unavailable due to privacy concerns.
Starting with the EU's GDPR, privacy protection legislation has been developed globally, and creating large-scale datasets requires satisfying the privacy policy under the law.
This paper explores an alternative SSL approach without collecting large-scale unlabeled datasets.

\subsection{Generative Semi-supervised Learning}\label{sec:gssl}
Generative semi-supervised learning (gSSL) is a variant of SSL where \(\mathcal{D}_\mathrm{u}\) is prohibited from being accessed and the unlabeled data \(x_\mathrm{u}\) is provided by a generative foundation model \(G_\mathrm{F}\) by
\begin{align}\label{eq:gssl_sample}
    x_\mathrm{u} = G_\mathrm{F}(z,\hat{y}_\mathrm{F}),
\end{align}
where \(\hat{y}_\mathrm{F}\) is an estimated foundation label produced by gSSL algorithm.
The gSSL algorithms have been rarely studied except for a prior work by \cite{Yamaguchi_arXiv22_PSSL}.
In a transfer learning setting where the target and source architectures are not consistent, \cite{Yamaguchi_arXiv22_PSSL} have proposed a method called pseudo semi-supervised learning (P-SSL).
Although P-SSL is focused on transfer learning, we consider it a simple baseline of gSSL.
P-SSL trains \(f_\theta\) by using Eq.~(\ref{eq:ssl_loss}) and estimates a foundation label \(\hat{y}_\mathrm{F}\) as
\begin{align}\label{eq:pssl_gen}
    \hat{y}_\mathrm{F} = f_{\theta_\mathrm{F}}(x),
\end{align}
where \(f_{\theta_\mathrm{F}}\) is a classifier pre-trained on a foundation dataset (e.g., an ImageNet pre-trained classifier).
That is, P-SSL interprets the training sample \(x\) as the conditional sample of an interpolated class in \(\mathcal{Y}_\mathrm{F}\) through the output of \(f_{\theta_\mathrm{F}}\).
This assumes the existence of \(f_{\theta_\mathrm{F}}\) and \(y\in\mathcal{Y}\) can be semantically approximated by the soft foundation labels, i.e., \(y_i \in \mathcal{Y} \approx f_{\theta_\mathrm{F}}(x_i) \in \mathcal{Y}_\mathrm{F}\).
However, the synthetic samples by Eq.~(\ref{eq:pssl_gen}) do not always contribute to the performance of \(f_\theta\) because the above assumption does not necessarily hold, and the synthetic samples are not directly optimized to improve \(f_\theta\).
In fact, \cite{Yamaguchi_arXiv22_PSSL} have reported that the performance gain by P-SSL is limited when the training datasets are not well approximated by Eq.~(\ref{eq:pssl_gen}).
To stably improve the performance of \(f_\theta\), we present a meta-learning based SSL approach, which does not require the label assumptions.

\begin{figure}[t]
    \centering
    \includegraphics[width=1.0\columnwidth]{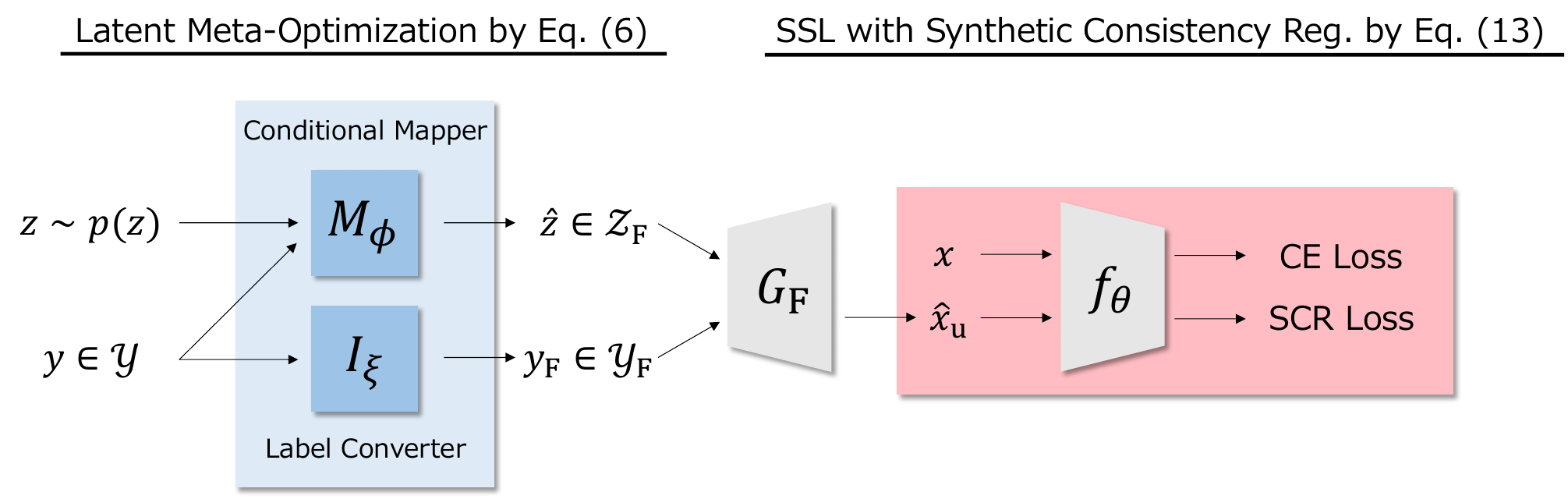}
    \caption{
     Overview of MP-SSL. We first generate a transformed latent variable \(\hat{z}\) and a pseudo foundation label \(y_\mathrm{F}\) through conditional mapper \(M_\phi\) and label converter \(I_\xi\). Then, we produce a pseudo unsupervised sample \(\hat{x}_\mathrm{u}=G_\mathrm{F}(\hat{z}, \hat{y}_\mathrm{F})\) for semi-supervised learning (SSL) of \(f_\theta\). To find the optimal \(\hat{z}\) and \(y_\mathrm{F}\), we update \(M_\phi\) and \(I_\xi\) by latent meta-optimization (LMO, Eq.~(\ref{eq:lmo})). In the training of \(f_\theta\), we use the loss of synthetic consistency regularization (SCR, Eq.~(\ref{eq:scr})) instead of existing SSL loss terms.
    }
    \label{fig:mpssl}
\end{figure}

\section{Proposed Method}
In this section, we describe our proposed method called MP-SSL.
MP-SSL is composed of (i) latent meta-optimization (LMO) and (ii) synthetic consistency regularization (SCR).
LMO finds synthetic samples performing as unlabeled data in SSL through meta-optimizing input latent variables and foundation class labels.
SCR penalizes a feature extractor by maximizing the similarity between variations of a synthetic sample.
MP-SSL alternately updates the parameters for sampling synthetic unlabeled data by LMO and training model \(f_\theta\) by SRC.
The overview of MP-SSL is illustrated in Figure~\ref{fig:mpssl}.

\subsection{Latent Meta-Optimization (LMO)}\label{sec:lmo}
The goal of LMO is to find a synthetic sample that approximates unlabeled data and contributes to the performance of \(f_\theta\) through SSL.
To extract unlabeled samples from \(G_\mathrm{F}\), we optimize the parameters \(\phi\) and \(\xi\) that generate the latent variables \(\hat{z}\in\mathcal{Z}_\mathrm{F}\) and foundation class label \(\hat{y}_\mathrm{F}\in\mathcal{Y}_\mathrm{F}\), respectively.
That is, we search for an optimal pair of \((\hat{z}, \hat{y}_\mathrm{F})\) through this optimization process.
In conditional generative models, the latent variables control overall characteristics without class categories (e.g., size of object and style), and the class labels determine the category of the synthetic samples~\citep{odena_ICML17_acgan,brock2018biggan}.
Searching \((\hat{z}, \hat{y}_\mathrm{F})\) can be more reasonable than directly optimizing whole parameters of \(G_\mathrm{F}\) on \(\mathcal{D}\) because the latter suffers from overfitting and the low-performance of \(f_\theta\) due to the low-quality samples~\citep{karras_NeurIPS20_ada}.
For the optimization, we use specialized architectures called \textit{conditional mapper} \(M_\phi:\mathcal{Z}_\mathrm{F}\times\mathcal{Y}\to\mathcal{Z}_\mathrm{F}\) and \textit{label converter} \(I_\xi:\mathcal{Y}\to\mathcal{Y}_\mathrm{F}\).
Through optimizing \(M_\phi\) and \(I_\xi\), we seek a synthetic sample \(\hat{x}_\mathrm{u}=G_\Phi(M_\phi(z_\mathrm{F},y),I(y))\).
To this end, we formalize the optimization problem of LMO as follows.
\begin{eqnarray}\label{eq:lmo}
    &\min\limits_{\phi,\xi} \mathcal{L}_\text{val}(\theta^*) + \lambda_\text{gap} \mathcal{L}_\text{gap}(\phi, \xi)\\
    &\mathcal{L}_\text{val}(\theta^*) = \mathbb{E}_{(x_\text{val}, y_\text{val})\in\mathcal{D}_\text{val}}\ell(f_{\theta^*}(x_\text{val}),y_\text{val})\\
    &\mathcal{L}_\text{gap}(\phi,\xi) = \mathbb{E}_{x\in\mathcal{D}}\|g_\psi(x) - g_\psi(\hat{x}_\mathrm{u}=G_\mathrm{F}(M_\phi(z,y),I_\xi(y)))\|^2_2\\
    \text{s.t.} & \theta^* = \argmin_\theta \mathcal{L}(\theta)+\lambda\mathcal{L}_\mathrm{u}(\theta, \phi,\xi),
\end{eqnarray}
where \(\mathcal{L}_\text{val}\) is for seeking samples to improve \(f_\theta\) and \(\mathcal{L}_\text{gap}\) is for satisfying that \(\hat{x}_\mathrm{u}\) approximates training data \(x\) as unlabeled samples.
This meta-optimization problem can be solved by stochastic gradient descent by extending the prior meta-learning method such as MAML~\citep{Finn_ICML17_maml}.
In the rest of this subsection, we describe the design of the conditional mapper and label converter.

\paragraph{Conditional Mapper \(M_\phi\).}
The role of \(M_\phi\) is to find optimal latent variables producing useful unlabeled samples for SSL through \(G_\mathrm{F}\).
Our idea is to transform the concatenation input latent variable \(z\) and training class label \(y\) into a new latent variable \(\hat{z}\).
This is based on an expectation that partitioning the problem for each class will make searching latent variables easier; we confirm that using \(y\) yields more performance gain in Sec.~\ref{sec:eval_abl_M}.
\(M_\phi\) outputs the estimated latent variable \(\hat{z}\) by
\begin{eqnarray}\label{eq:cond_mapper}
    \hat{z} = M_\phi(z,y) = \operatorname{MLP}_\phi(\operatorname{Concat}(z, \operatorname{EMB}_\phi(y))),
\end{eqnarray}
where \(\operatorname{EMB}_\phi:\mathcal{Y}\to\mathbb{R}^{d_\mathcal{Y}}\) is an embedding layer for \(y\), \(\operatorname{Concat}(\cdot)\) is a concatenation operation of two vectors, and \(\operatorname{MLP}_\phi:\mathbb{R}^{d_{\mathcal{Z}_\mathrm{F}}+d_\mathcal{Y}}\to\mathcal{Z}_\mathrm{F}=\mathbb{R}^{d_{\mathcal{Z}_\mathrm{F}}}\) is a multi-layer perception yielding a new latent variable.

\paragraph{Label Converter \(I_\xi\).}
\(I_\xi\) estimates a foundation label \(\hat{y}_\mathrm{F}\) corresponding to a training class label \(y\).
To estimate a foundation label, a prior work~\citep{Yamaguchi_arXiv22_PSSL} utilizes a pre-trained classifier on foundation datasets.
This approach is simple, but the pre-trained classifiers are not necessarily given, and the estimation of foundation soft labels depends on the performance of the pre-trained classifiers.
Thus, if high-performance pre-trained classifiers are unavailable, it is hard to estimate a foundation label correctly.
Instead of the pre-trained classifiers, we utilize the Gumbel-softmax~\citep{Jang_ICLR17_gumbel_softmax} trick for sampling \(\hat{y}_\mathrm{F}\) through the parameter \(\xi\) updated by LMO:
\begin{eqnarray}\label{eq:label_converter}
    &\hat{y}_\mathrm{F} = \argmax_i~I_\xi(y),\\
    &I_\xi(y)[i] = \frac{\exp{((\log (\operatorname{EMB}_\xi[i])+\mathbf{g}[i])/\tau)}}{\sum^{|\mathcal{Y}_\mathrm{F}|}_{j=1}\exp{((\log (\operatorname{EMB}_\xi[j])+\mathbf{g}[j])/\tau)}},
\end{eqnarray}
where \(\operatorname{EMB}_\xi:\mathcal{Y}\to\mathbb{R}^{d_\mathcal{Y}}\) is an embedding layer for \(y\), \(\mathbf{g}[i] = -\log(-\log (u_i\sim \operatorname{Uniform}(0,1)))\), \(\tau\) is a temperature parameter.
This formulation has several advantages: (a) it can be trained by backpropagation since it is fully differentiable, (b) the output \(\hat{x}_\mathrm{u}\) is expected to be unbiased due to randomness given by \(\mathbf{g}\), and (c) the number of foundation classes of interest can be adjustable according to the training data by the temperature parameters.
We confirm these advantages through comparison to the other variants of \(I_\xi\) in Sec.~\ref{sec:eval_abl_I}.

\begin{algorithm}[t]
    \caption{MP-SSL}\label{alg:mpssl}
    \begin{algorithmic}[1]
    {\small
        \REQUIRE{Training dataset \(\mathcal{D}\), validation dataset \(\mathcal{D}_\text{val}\)  classifier \(f_\theta\), generative foundation model \(G_\mathrm{F}\), conditional mapper \(M_\phi\), label converter \(I_\xi\), training batchsize \(B\), validation batchsize \(B_\text{val}\), step size \(\eta\) and \(\xi\), hyperparameter \(\lambda\)
        \ENSURE{Trained classifier \(f_\theta\)}}
        \WHILE{not converged}
        \STATE{\(\{(x^i,y^i)\}^B_{i=1}\sim \mathcal{D}\)}
        \STATE{\(\{z^i\}^{B}_{i=1} \sim \mathcal{N}(0,I)\)}
        \STATE{// Updating \(\phi\) and \(\xi\) by LMO}
        \STATE{\(\{(x^i_\text{val},y^i_\text{val})\}^{B_\text{val}}_{i=1}\sim \mathcal{D}\)}
        \STATE{\(\{\hat{x}_\mathrm{u}^i\}^{B}_{i=1} = \{G_\mathrm{F}(M_\phi(z^i, y^i),I_\xi(y^i)\}^{B}_{i=1}\)}
        \STATE{\(\theta^\prime \leftarrow \theta - \eta \nabla_\theta(\frac{1}{B}\sum_{i=1}^{B}\ell(f_\theta(x^i),y^i) + \frac{\lambda}{B}\sum_{i=1}^{B}\ell_\text{SCR}(\hat{x}_\mathrm{u}^i;\psi))\)}
        \STATE{\(\phi \leftarrow \phi - \xi \nabla_\phi{(\frac{1}{B_\text{val}}\ell(f_{\theta^\prime}(x_\text{val}), y_\text{val}) + \|\frac{1}{B}\sum_{i=1}^{B}f_{\theta}(x^i) - \frac{1}{B}\sum_{i=1}^{B}f_{\theta}(\hat{x}_\mathrm{u}^i)\|^2_2)}\)}
        \STATE{// Updating \(\theta\) with SCR}
        \STATE{\(\{\hat{x}_\mathrm{u}^i\}^{B}_{i=1} = \{G_\Phi(F_\phi(z^i),y_\mathrm{p}^i)\}^{B}_{i=1}\)}
        \STATE{\(\theta \leftarrow \theta - \eta \nabla_\theta(\frac{1}{B}\sum_{i=1}^{B}\ell(f_\theta(x^i),y^i) + \frac{\lambda}{B}\sum_{i=1}^{B}\ell_\text{SCR}(\hat{x}_\mathrm{u}^i;\psi))\)}
        \ENDWHILE
        }
    \end{algorithmic}
\end{algorithm}

\subsection{Synthetic Consistency Regularization}\label{sec:scr}
Although synthetic samples generated from \(G_\mathrm{F}\) through LMO can contain useful information for training \(f_\theta\), it is hard to expect that they are exactly categorized to the training space \(\mathcal{Y}\) because the training and foundation label spaces are not the same, i.e., \(\mathcal{Y}\neq\mathcal{Y}_\mathrm{F}\).
Therefore, training with the synthetic samples via unsupervised losses using pseudo training labels in \(\mathcal{Y}\) (e.g., FixMatch~\citep{sohn_NIPS20_fixmatch}) might confuse \(f_\theta\) due to the label space mismatch.
To avoid the negative effect and maximize the gain from the synthetic samples, we introduce a simple unsupervised loss called synthetic consistency regularization (SCR).
In contrast to existing pseudo-label based SSL methods, SCR is computed on the feature extractor \(g_\psi\) of \(f_\theta\).
That is, we regularize \(g_\psi\) by synthetic samples instead of the classifier head \(h_\omega\).
To regularize \(g_\psi\), we design SCR based on consistency regularization~\citep{bachman_NIPS14_learning_with_pseudo_ensembles,xie_NIPS20_UDA,sohn_NIPS20_fixmatch}, which minimizes the gap between the outputs of two variants of samples that are transformed by different data augmentations.
Concretely, we formalize the loss function of SCR as follows.
\begin{eqnarray}\label{eq:scr}
\ell_\text{SCR}(\hat{x}_\mathrm{u};\psi) &=& 1 - \frac{g_\psi(T_\mathrm{w}(\hat{x}_\mathrm{u}))\cdot g_\psi(T_\mathrm{s}(\hat{x}_\mathrm{u}))}{\|g_\psi(T_\mathrm{w}(\hat{x}_\mathrm{u}))\|\|g_\psi(T_\mathrm{s}(\hat{x}_\mathrm{u}))\|}, 
\end{eqnarray}
where \(T_\mathrm{w}(\cdot)\) and \(T_\mathrm{s}(\cdot)\) are a weak augmentation (e.g., flip and crop) and a strong augmentation (e.g., RandAugment~\citep{cubuk_CVPR20_randaugment}).
As the measurement of the gap, we choose cosine distance; we empirically found that this formulation achieves the best results when comparing with L2, L1, and smooth L1 distance as shown in Sec.~\ref{sec:eval_abl_SCR}.
By applying SCR to \(g_\psi\), we expect that \(g_\psi\) learns robust feature representations that are useful for classifying real samples by \(h_\omega\).

Finally, we show the overall procedure of MP-SSL using LMO and SCR in Algorithm~\ref{alg:mpssl}.

\section{Experiments}~\label{sec:experiment}
This section evaluates our MP-SSL through experiments on multiple image classification datasets.
We mainly aim to answer three research questions with the experiments:
(1) Can MP-SSL improve the baselines without real unlabeled datasets?
(2) What can training models learn through MP-SSL?
(3) Is the MP-SSL design reasonable?
We compare MP-SSL with baselines with synthetic samples, e.g., P-SSL~\citep{Yamaguchi_arXiv22_PSSL}, and baselines with real samples e.g., FreeMatch~\citep{Wang_ICLR23_freematch} in Sec.~\ref{sec:eval_multiple_dataset}~and~\ref{sec:eval_dataset_size}.
Furthermore, we provide a detailed analysis of MP-SSL, such as the visualization of synthetic samples (Sec.~\ref{sec:eval_synthetic_samples}) and detailed ablation studies of MP-SSL (Sec.~\ref{sec:eval_ablation}).

\subsection{Setting}
\paragraph*{Baselines.}
We compare our method with the following baselines in the gSSL setting.
\textbf{Base Model}: training \(f_\theta\) with only \(\mathcal{D}\).
\textbf{Na\"ive gSSL}: training $f_\theta$ with $\mathcal{D}$ and $G_\mathrm{F}$, where a synthetic sample $\hat{x_\mathrm{u}}$ is generated from uniformly sampled $z$ and $y_\mathrm{F}$, then we train $f_\theta$ by an existing SSL method with the real and synthetic samples.
\textbf{P-SSL}~\citep{Yamaguchi_arXiv22_PSSL}: training \(f_\theta\) with \(\mathcal{D}\) and \(G_\mathrm{F}\) with sampling \(y_\mathrm{F}\) by Eq.~(\ref{eq:pssl_gen}) and existing SSL methods updating \(h_\omega\).
We also test SSL methods using a real unlabeled dataset \(\mathcal{D}_\mathrm{u}\) to assess the practicality of the gSSL setting; We refer this setting to oracle SSL because they can access \(\mathcal{D}_\mathrm{u}\) that is prohibited in gSSL.
As the oracle SSL methods, We used three representative SSL methods: UDA~\citep{xie_NIPS20_UDA}, FixMatch~\citep{sohn_NIPS20_fixmatch}, and FreeMatch~\citep{Wang_ICLR23_freematch}.

\paragraph*{Datasets.}
We used six image datasets for classification tasks: Cars~\citep{krause_3DRR2013_stanford_cars}, Aircraft~\citep{maji_13_aircraft}, Birds~\citep{Welinder_10_cub2002011}, DTD~\citep{cimpoi_CVPR14_DTD}, Flowers~\citep{Nilsback_08_flowers}, and Pets~\citep{parkhi_CVPR12_oxford_pets}.
To evaluate both generative and oracle SSL settings at the same time, we randomly split them into \(\mathcal{D}\) and \(\mathcal{D}_\mathrm{u}\) (\(50:50\), by default), and discarded \(\mathcal{D}_\mathrm{u}\) in gSSL and used in oracle SSL.
Furthermore, to evaluate the effect of dataset size, we varied the size of labeled datasets of Cars by \(\{10,25,50,100\}\%\) in volume.
Note that we used all of the rest of the unlabeled samples as \(\mathcal{D}_\mathrm{u}\) in this setting.
After creating \(\mathcal{D}\), we randomly split \(\mathcal{D}\) into \(9:1\) and used the former as \(\mathcal{D}\) and the latter as \(\mathcal{D}_\text{val}\) in the training.

\paragraph*{Architectures.}
We used ResNet-18~\citep{he_resnet} as \(f_\theta\) and BigGAN for \(256\times256\) images~\citep{brock2018biggan} as \(G_\mathrm{F}\).
\(M_\phi\) was composed of a three-layer perceptron with a leaky-ReLU activation function.
We used the ImageNet pre-trained weights of ResNet-18 distributed by PyTorch.\footnote{https://github.com/pytorch/vision}
For BigGAN, we used the ImageNet pre-trained weights provided by \cite{brock2018biggan}. 
Note that we used the same \(G_\mathrm{F}\) in the baselines and our method.\looseness=-1

\paragraph*{Training.}
We trained \(f_\theta\) by the Nesterov momentum SGD for 200 epochs with a momentum of 0.9 and an initial learning rate of 0.01; we decayed the learning rate by 0.1 at 60, 120, and 160 epochs.
We trained \(M_\phi\) and \(I_\xi\) by the Adam optimizer for 200 epochs with a learning rate of \(1.0\times10^{-4}\).
We used mini-batch sizes of 64.
The input samples were resized into a resolution of \(224\times224\); \(\hat{x}_\mathrm{u}\) was resized by differentiable transformations. For synthetic samples from \(G_\mathrm{F}\) in MP-SSL, the weak transformation \(T_\mathrm{w}\) was horizontal flip and random crop, and the strong transformation \(T_\mathrm{s}\) was RandAugment~\citep{cubuk_CVPR20_randaugment} by following~\cite{xie_NIPS20_UDA}; it was implemented with differentiable transformations provided in Kornia~\citep{Riba_WACV20_kornia}.
We determined the hyperparameter \(\lambda\) by grid search among \([0.1,1.0]\) with a step size of \(0.1\) for each method by \(\mathcal{D}_\text{val}\).
We used \(\lambda_\text{gap}\) of \(10\).
For the hyperparameters of oracle SSL methods, we followed the default settings of the original papers~\citep{xie_NIPS20_UDA,sohn_NIPS20_fixmatch,Wang_ICLR23_freematch}.
We selected the final model by checking the validation accuracy for each epoch.
We ran the experiments three times on a 24-core Intel Xeon CPU with an NVIDIA A100 GPU with 40GB VRAM and recorded average test accuracies with standard deviations evaluated on the final models.

\begin{table}[t]
    \centering
        \caption{
            Performance comparison of ResNet-18 classifiers on multiple datasets (Top-1 Acc. (\%)).
            \underline{Underlined scores} are the best of the oracle SSL setting (i.e., using real unlabeled datasets), and \textbf{Bolded scores} are the best among the methods of the generative SSL (gSSL) setting (i.e., using foundation generative models).
            }
        \label{tb:multiple_dataset}
        \resizebox{0.9\columnwidth}{!}{
            \begin{tabular}{lccccccccc}\toprule
                Method / Dataset & Aircraft & Birds & Cars & DTD & Flower & Pets  \\
              \midrule
                Base Model    &  44.05\(^{\pm\text{.59}}\) & 60.74\(^{\pm\text{.29}}\) &  71.62\(^{\pm\text{.30}}\) & 61.56\(^{\pm\text{.56}}\) & 88.14\(^{\pm\text{.18}}\) & 84.44\(^{\pm\text{.48}}\) \\ \midrule
                \textbf{Oracle SSL (\(\mathcal{D} + \mathcal{D}_\mathrm{u}\))}\\
                UDA~\citep{xie_NIPS20_UDA} &  44.65\(^{\pm\text{.38}}\) & 60.22\(^{\pm\text{.03}}\) &  60.22\(^{\pm\text{.03}}\) & 70.90\(^{\pm\text{.58}}\) & 61.90\(^{\pm\text{.10}}\) & \underline{87.72}\(^{\pm\text{.31}}\) \\
                FixMatch~\citep{sohn_NIPS20_fixmatch} &  47.89\(^{\pm\text{.38}}\) & 60.58\(^{\pm\text{.84}}\) &  80.98\(^{\pm\text{.36}}\) & 61.31\(^{\pm\text{.11}}\) & \underline{90.08}\(^{\pm\text{.48}}\) & 81.73\(^{\pm\text{.39}}\) \\
                FreeMatch~\citep{Wang_ICLR23_freematch} &  \underline{49.55}\(^{\pm\text{.33}}\) & \underline{66.09}\(^{\pm\text{.16}}\) & \underline{82.73}\(^{\pm\text{.41}}\) & \underline{63.83}\(^{\pm\text{.49}}\) & 90.07\(^{\pm\text{.27}}\) & 86.61\(^{\pm\text{.40}}\) \\\midrule
                \textbf{Generative SSL (\(\mathcal{D} + G_\mathrm{F}\))}\\
                Na\"ive gSSL (FreeMatch) &  46.83\(^{\pm\text{.34}}\) & 60.95\(^{\pm\text{.29}}\) &  73.67\(^{\pm\text{.67}}\) & 59.41\(^{\pm\text{.17}}\) & 86.41\(^{\pm\text{.25}}\) & 83.66\(^{\pm\text{.69}}\) \\
                P-SSL~\citep{Yamaguchi_arXiv22_PSSL}   &  45.43\(^{\pm\text{.24}}\) &  60.54\(^{\pm\text{.25}}\) & 72.45\(^{\pm\text{.30}}\) & 60.82\(^{\pm\text{.61}}\) & 88.20\(^{\pm\text{.15}}\) & 84.84\(^{\pm\text{.41}}\)\\
                MP-SSL (Ours) &  \textbf{49.48}\(^{\pm\textbf{.25}}\) & \textbf{62.86}\(^{\pm\textbf{.23}}\) &  \textbf{76.33}\(^{\pm\textbf{.31}}\) & \textbf{62.34}\(^{\pm\textbf{.46}}\) & \textbf{88.44}\(^{\pm\textbf{.51}}\) & \textbf{85.43}\(^{\pm\textbf{.09}}\) \\
                \bottomrule
            \end{tabular}
        }
        \vspace{-3mm}
 \end{table}

\subsection{Evaluation on Multiple Datasets}\label{sec:eval_multiple_dataset}
First, we evaluate our MP-SSL's performance by comparing it with the baseline methods of gSSL and oracle SSL on various training datasets.
Table~\ref{tb:multiple_dataset} shows the results on six datasets.
Note that we did not use the unlabeled dataset \(\mathcal{D}_\mathrm{u}\) in the gSSL setting.
Our MP-SSL achieved the best results among the gSSL methods with a large margin (up to 3pp).
While P-SSL degraded the base model on DTD due to the mismatch between training and foundation label spaces~\citep{Yamaguchi_arXiv22_PSSL}, our MP-SSL succeeded in improving it.
This indicates that MP-SSL is not sensitive to the label space mismatch and stably improves classifiers in various settings.
Furthermore, on the Aircraft and DTD datasets, MP-SSL is competitive with the oracle SSL methods.
This suggests that MP-SSL and gSSL have the potential to approximate the oracle SSL methods in terms of the final model accuracy.

\begin{table}[]
    \centering
    \centering
        \caption{
            Performance comparison of ResNet-18 classifiers on the reduced Cars datasets (Top-1 Acc. (\%)).
            \underline{Underlined scores} are the best of the oracle SSL setting (i.e., using real unlabeled datasets), and \textbf{Bolded scores} are the best among the methods of the gSSL setting (i.e., using foundation generative models).
        }\label{tb:dataset_size}
        \resizebox{0.7\columnwidth}{!}{
        \begin{tabular}{lcccc}\toprule
            Method / Labeled Dataset Size & 10\% & 25\% & 50\% & 100\% \\\midrule
            Base Model   &  19.74\(^{\pm\text{.15}}\) &47.54\(^{\pm\text{.67}}\) &  71.62\(^{\pm\text{.30}}\) & 85.75\(^{\pm\text{.08}}\)\\\midrule
            \textbf{Oracle SSL (\(\mathcal{D} + \mathcal{D}_\mathrm{u}\))}\\
            UDA~\citep{xie_NIPS20_UDA}   &  19.36\(^{\pm\text{.44}}\) &47.95\(^{\pm\text{.30}}\) &  72.76\(^{\pm\text{.53}}\) & N/A\\
            FixMatch~\citep{sohn_NIPS20_fixmatch}   &  \underline{20.98}\(^{\pm\text{.99}}\) &\underline{63.58}\(^{\pm\text{.64}}\) &  \underline{83.94}\(^{\pm\text{.65}}\) & N/A\\
            FreeMatch~\citep{Wang_ICLR23_freematch}   &  18.07\(^{\pm\text{.03}}\) &60.13\(^{\pm\text{.61}}\) &  82.60\(^{\pm\text{.28}}\) & N/A \\\midrule
            {\textbf{Generative SSL (\(\mathcal{D} + G_\mathrm{F}\))}}\\
            Na\"ive gSSL (FreeMatch)  &  20.11\(^{\pm\text{.03}}\) &49.33\(^{\pm\text{.54}}\) &  72.91\(^{\pm\text{.38}}\) & 81.68\(^{\pm\text{.18}}\)\\
            P-SSL~\citep{Yamaguchi_arXiv22_PSSL}   &  20.34\(^{\pm\text{.42}}\) &48.27\(^{\pm\text{.48}}\) &  72.62\(^{\pm\text{.33}}\) & 85.78\(^{\pm\text{.23}}\)\\
            MP-SSL (Ours) &  {\bf 23.82}\(^{\pm\textbf{.55}}\) & {\bf 53.37}\(^{\pm\textbf{.56}}\) &  {\bf 76.33}\(^{\pm\text{.31}}\) & {\bf 86.84}\(^{\pm\textbf{.10}}\) \\
            \bottomrule
        \end{tabular}
        }
\end{table}

\begin{figure}[t]
    \centering
        \begin{minipage}{0.3\textwidth}
            \subfigure[Aircraft]{\includegraphics[width=\columnwidth]{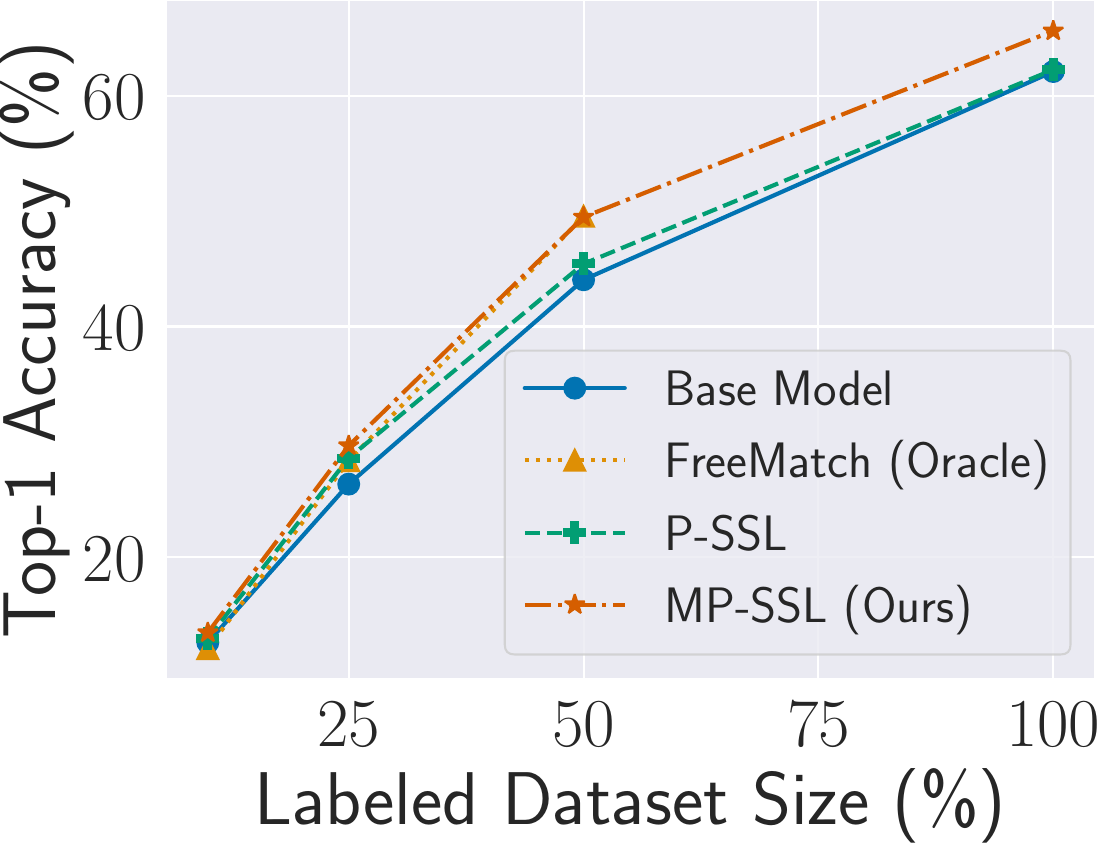}}\label{fig:aircraft_datasetsize}
        \end{minipage}
        \begin{minipage}{0.3\textwidth}
            \subfigure[Bird]{\includegraphics[width=\columnwidth]{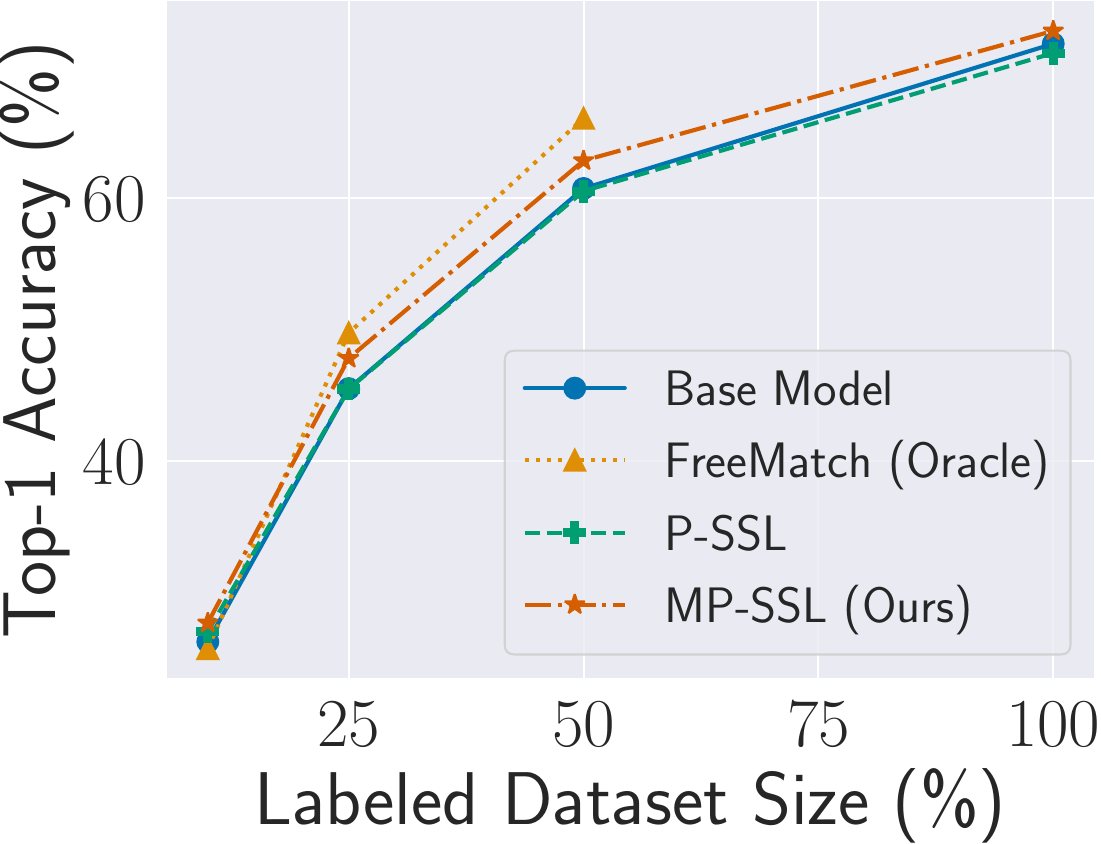}}\label{fig:bird_datasetsize}
        \end{minipage}
        \begin{minipage}{0.3\textwidth}
            \subfigure[Cars]{\includegraphics[width=\columnwidth]{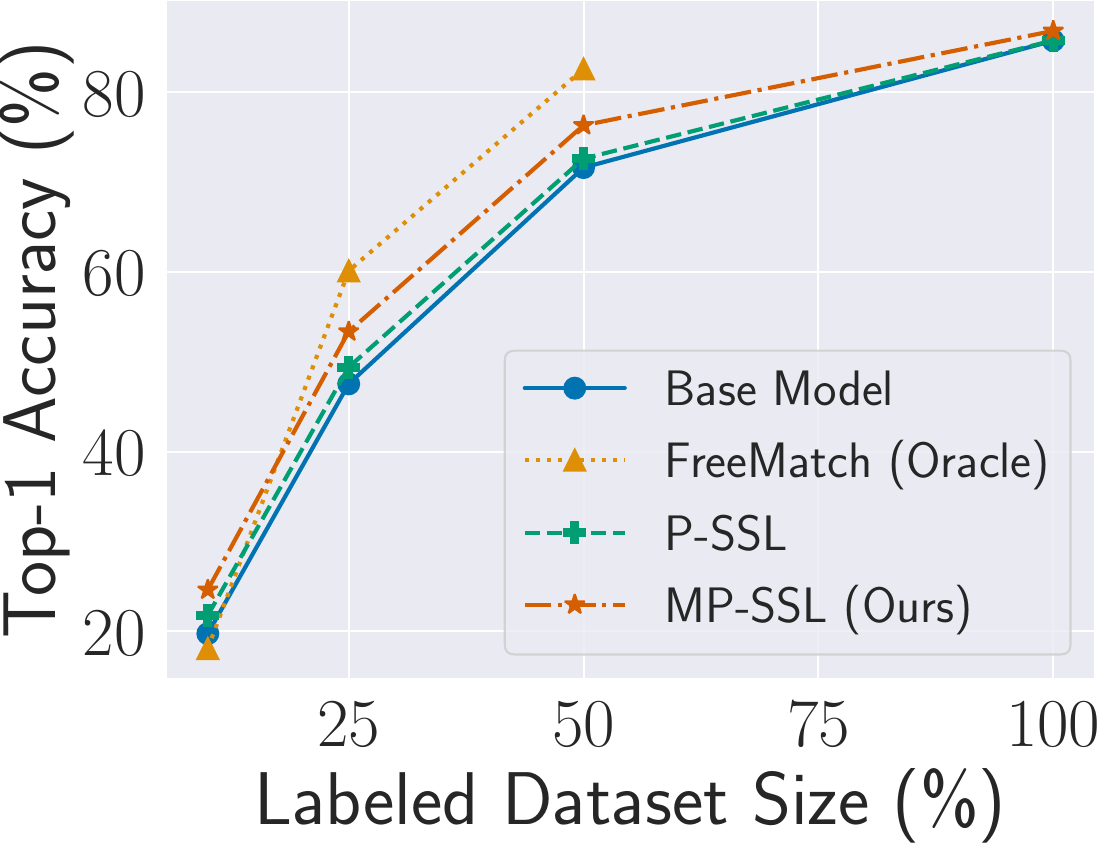}}\label{fig:cars_datasetsize}
        \end{minipage}
    \caption{Performance Comparisons in Small Labeled Dataset (ResNet-18)}
    \label{fig:dataset_size}
    \vspace{-3mm}
\end{figure}

\subsection{Evaluation by Varying Dataset Size}\label{sec:eval_dataset_size}
We evaluate MP-SSL by varying the size of training labeled datasets.
We used all of the remaining unlabeled samples as \(\mathcal{D}_\mathrm{u}\) for the oracle SSL methods and did not use \(\mathcal{D}_\mathrm{u}\) for the gSSL methods.
Table~\ref{tb:dataset_size} shows that our MP-SSL achieves the best results in the gSSL setting for all dataset sizes.
More interestingly, MP-SSL significantly outperformed the best result of the oracle SSL methods when the labeled dataset is extremely small (i.e., 10\% \(\leq\) 1,000 samples).
This trend is consistent with multiple datasets, as shown in Fig.~\ref{fig:dataset_size}.
These results suggest that the synthetic samples from \(G_\mathrm{F}\) are more valuable than real unlabeled samples for improving classification performance when the labeled datasets are quite small.

\begin{figure}[t]
    \centering
        \begin{minipage}{0.3\textwidth}
            \subfigure[Real]{\includegraphics[width=0.9\columnwidth]{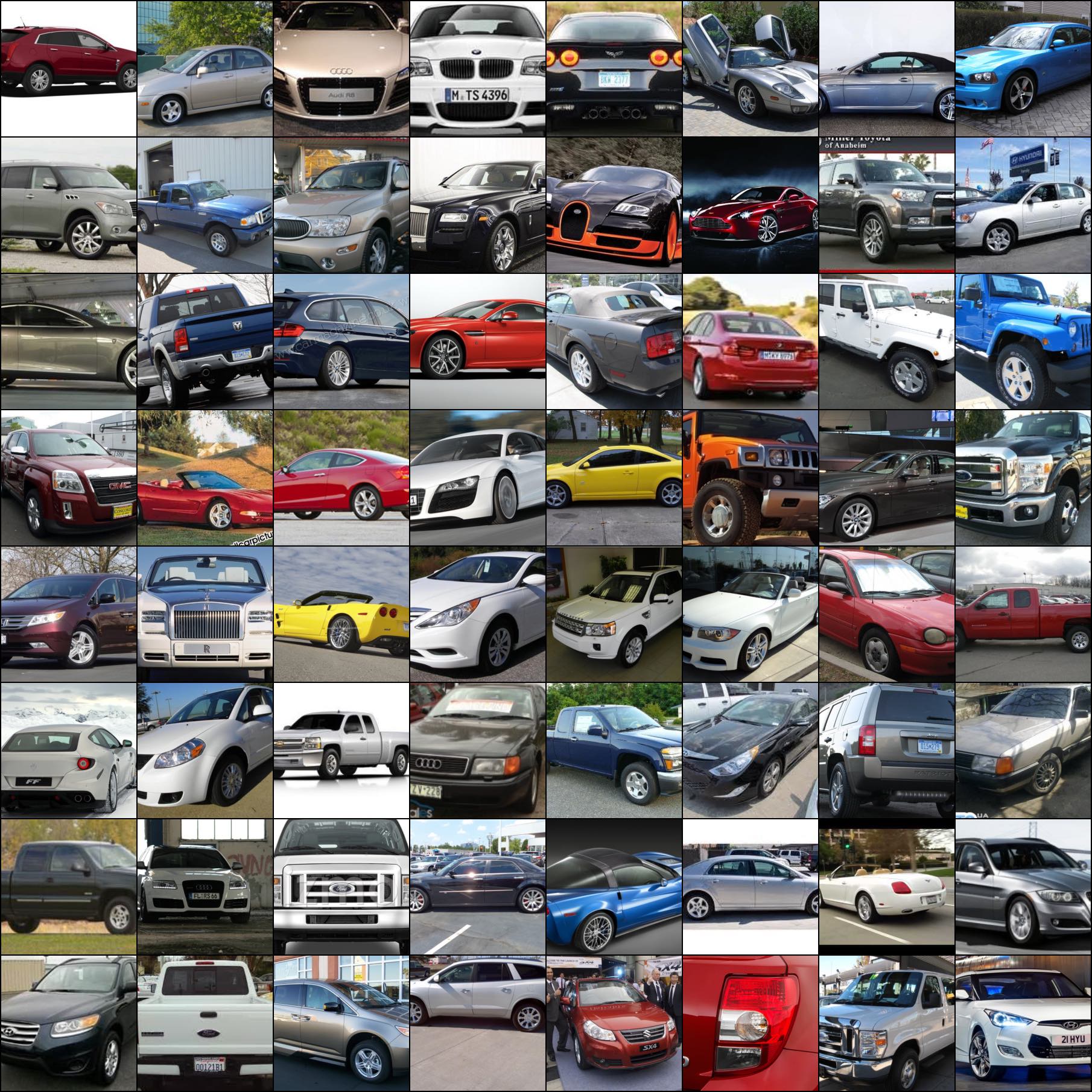}}\label{fig:real_sample}
        \end{minipage}
        \begin{minipage}{0.3\textwidth}
            \subfigure[P-SSL]{\includegraphics[width=0.9\columnwidth]{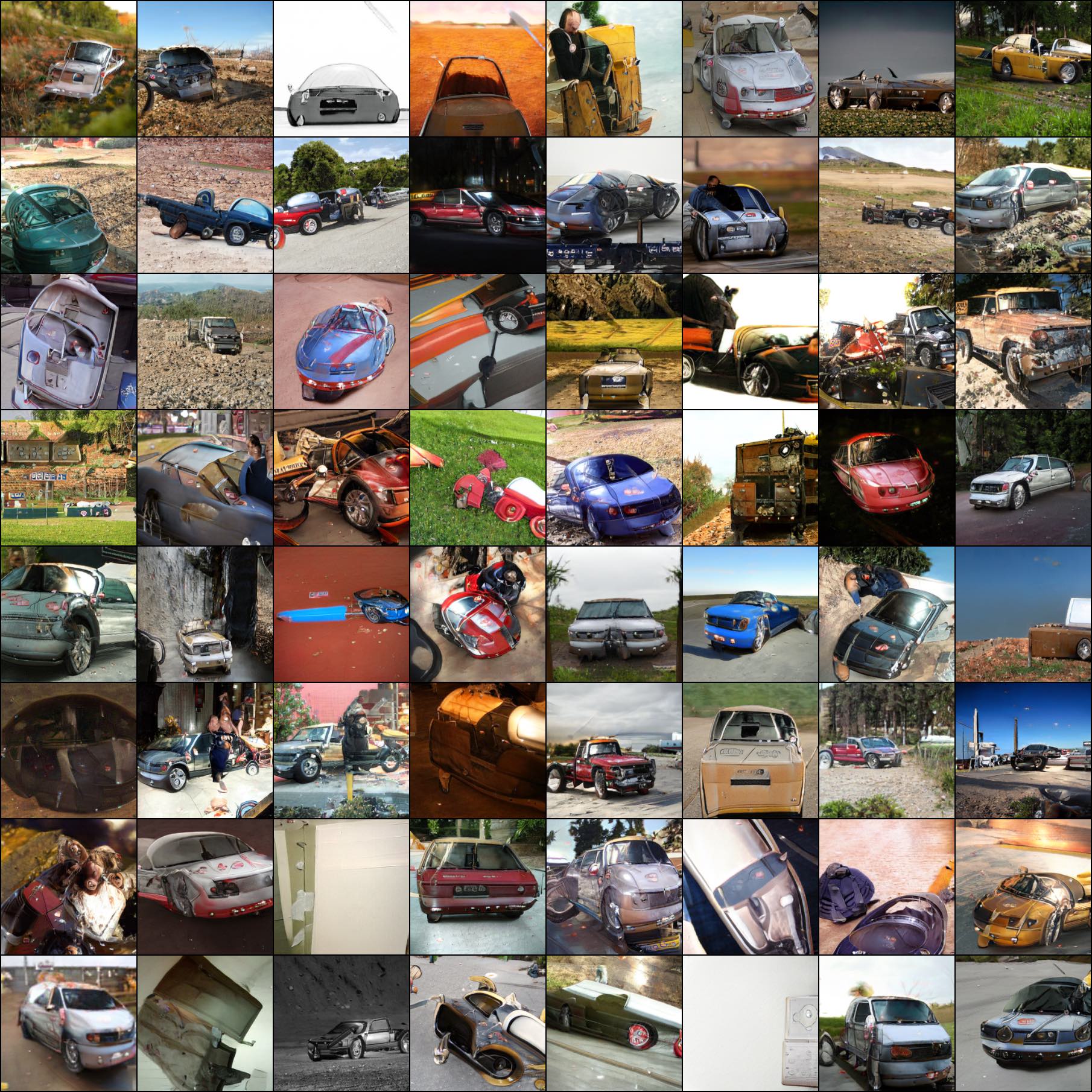}}\label{fig:pssl_sample}
        \end{minipage}
        \begin{minipage}{0.3\textwidth}
            \subfigure[MP-SSL (Ours)]{\includegraphics[width=0.9\columnwidth]{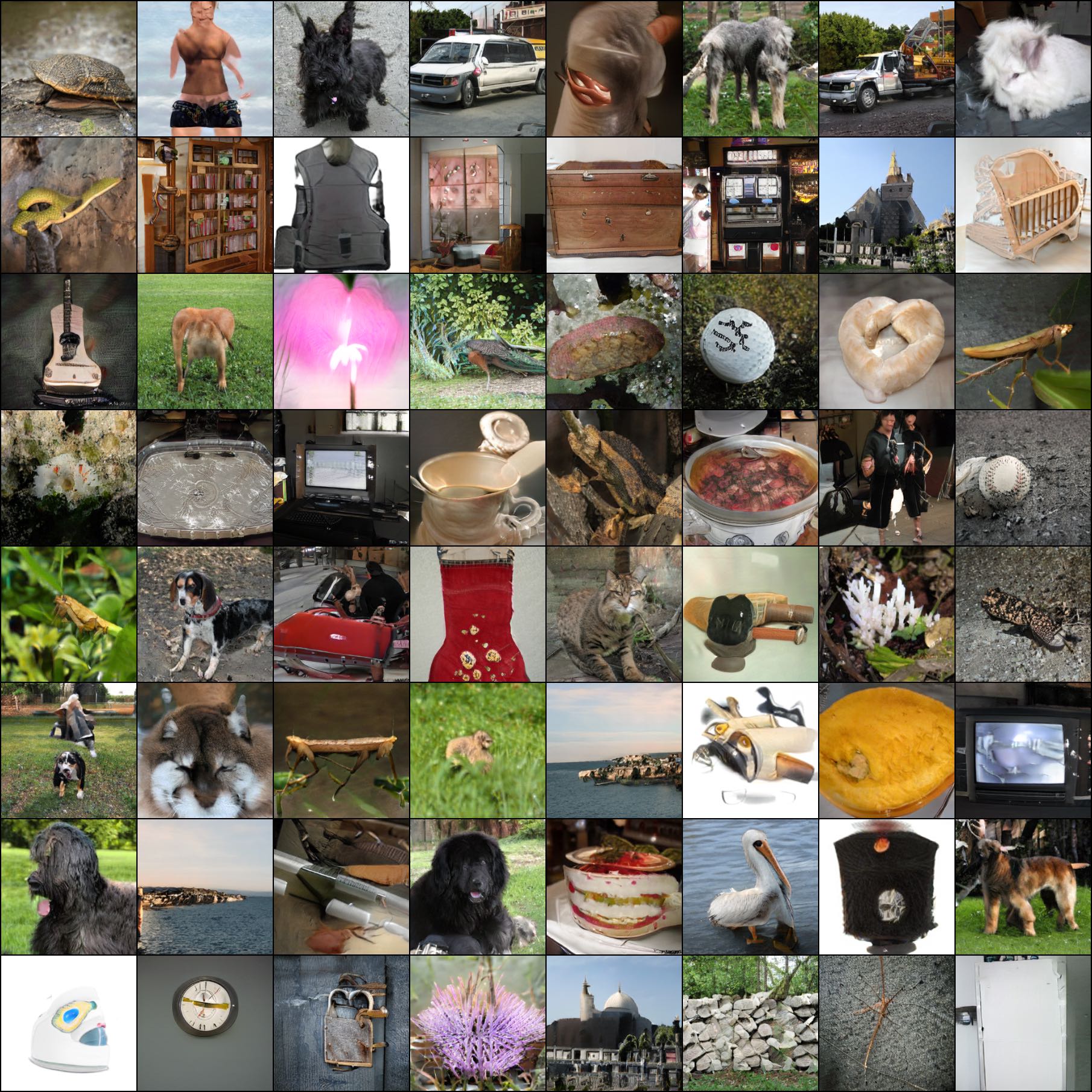}}\label{fig:mpssl_sample}
        \end{minipage}
    \caption{Real and Synthetic Samples in Training (Cars)}
    \label{fig:generated_samples}
\end{figure}

\subsection{Analysis of Synthetic Samples}\label{sec:eval_synthetic_samples}
We examine what the classifier is learning through MP-SSL.
To this end, we visualize the synthetic samples generated by MP-SSL and compare them to real and synthetic samples generated by P-SSL.
Figure~\ref{fig:generated_samples} shows the real and synthetic samples.
Interestingly, we see that P-SSL produces more relative samples to real samples (Cars), whereas MP-SSL produces not so relative but diverse samples.
Since the performance studies in Sec.~\ref{sec:eval_multiple_dataset}~and~\ref{sec:eval_dataset_size} show that MP-SSL completely outperformed P-SSL, this visualization result is contrary to intuition.
We consider that this can be caused by the unsupervised regularization of MP-SSL, which penalizes the feature extractor instead of the entire model.
As defined in Eq.~(\ref{eq:lmo}), LMO of MP-SSL optimizes the latent vectors through the backpropagation from the unsupervised loss, and thus, the synthetic samples generated from the latent vectors are not optimized to become similar to real samples in its label spaces.
The results suggest that the regularization of feature extractors does not necessarily require perfect imitation of the training data, and the diversity of samples is more important.

\subsection{Ablation Study}\label{sec:eval_ablation}
\begin{table}[t]
    \centering
        \begin{minipage}{0.49\textwidth}
        \centering
            \caption{Analysis of LMO}
            \label{tb:abl_lmo}
            \resizebox{0.8\textwidth}{!}{
            \begin{tabular}{lc}\toprule
                 Pattern & Cars Test Acc.(\%)\\\midrule
                 Base Model & 71.62\(^{\pm\text{.30}}\) \\
                 MP-SSL w/o LMO & 74.34\(^{\pm\text{.01}}\) \\
                 MP-SSL w/o \(\mathcal{L}_\text{gap}(\phi, \xi)\) & 75.46\(^{\pm\text{.22}}\) \\
                 MP-SSL w/o \(\mathcal{L}_\text{val}(\theta^*)\) & 75.63\(^{\pm\text{.21}}\) \\
                 MP-SSL & \textbf{76.33}\(^{\pm\textbf{.31}}\) \\
            \end{tabular}
            }
        \end{minipage}
        \begin{minipage}{0.49\textwidth}
        \centering
            \caption{Ablation Study of \(M_\phi\)}
            \label{tb:abl_conditional_mapper}
            \resizebox{0.8\textwidth}{!}{
            \begin{tabular}{lc}\toprule
                 Pattern & Cars Test Acc.(\%)\\\midrule
                 Base Model & 71.62\(^{\pm\text{.30}}\) \\
                 Unconditional \(M_\phi\) & 75.55\(^{\pm\text{.40}}\) \\
                 Conditional \(M_\phi\) & \textbf{76.33}\(^{\pm\textbf{.31}}\) 
            \end{tabular}
            }
        \end{minipage}
        \vspace{-3mm}
\end{table}
\subsubsection{Meta-Learning and Gap Loss in LMO}\label{sec:eval_abl_LMO}
We evaluate the effectiveness of LMO by decomposing the objective function defined in Eq.~(\ref{eq:lmo}).
Eq.~(\ref{eq:lmo}) is composed of the meta-learning loss \(\mathcal{L}_\text{val}(\theta^*)\) and the feature gap loss \(\mathcal{L}_\text{gap}(\phi, \xi)\).
Table~\ref{tb:abl_lmo} shows the impact of these components on accuracy by ablating them in MP-SSL.
The row of MP-SSL w/o LMO denotes the test pattern of discarding LMO from MP-SSL, i.e., producing \(\hat{x}_\mathrm{u}\) by random sampling from \(G_\mathrm{F}\).
From the results, we confirm that \(\mathcal{L}_\text{val}(\theta^*)\) and \(\mathcal{L}_\text{gap}(\phi, \xi)\) equally contribute to the test performance.
In other words, the meta-learning loss and the feature gap loss have different effects on the synthetic samples and are complementary.

\subsubsection{Conditional Mapper}\label{sec:eval_abl_M}
We assess the design validity of conditional mapper \(M_\phi(z,y)\). In Eq.~(\ref{eq:cond_mapper}), we define \(M_\phi\) to be conditioned by a training class label \(y\).
To confirm the effectiveness of using labels, we tested unconditional mapper \(M_\phi(z)\), which is created by discarding the components for labels from \(M_\phi(z,y)\).
Table~\ref{tb:abl_conditional_mapper} summarises the results.
MP-SSL with a conditional mapper significantly outperformed one with an unconditional mapper.
Therefore, we can say that using conditional labels for transforming a latent variable \(z\) helps boost models' performance.\looseness-1

\subsubsection{Label Converter}\label{sec:eval_abl_I}
In Sec.~\ref{sec:lmo}, we design label converter \(I_\xi\) composed of the Gumbel softmax module as Eq.~(\ref{eq:label_converter}).
This section provides the ablation study to evaluate the design choice.
We varied the implementation of \(I_\xi\) with (a) soft label by embedding layer, i.e., \(\hat{y}_\mathrm{F} = \operatorname{EMB}_\xi\), (b) soft Gumbel softmax, i.e., \(\hat{y}_\mathrm{F} = I_\xi\).
Furthermore, we varied the hyperparameter \(\tau\) in Eq.~(\ref{eq:label_converter}).
Table~\ref{tb:abl_label_converter} shows the results.
Using the Gumbel softmax with hard label output brings better test accuracy.
This indicates that using the soft label output might not be appropriate for the unsupervised regularization loss since it results in ambiguous and low-quality output as in P-SSL, which uses soft labels for generating synthetic samples (Figure~\ref{fig:pssl_sample}).\looseness-1

\begin{table}[t]
    \centering
        \begin{minipage}{0.49\textwidth}
            \caption{Ablation Study of \(I_\xi\)}
            \label{tb:abl_label_converter}
            \resizebox{\textwidth}{!}{
            \begin{tabular}{lc}\toprule
                 Output Module & Cars Test Acc.(\%)\\\midrule
                 Soft Label by \(\operatorname{EMB}_\xi\) & 75.55\(^{\pm\text{.24}}\) \\
                 Soft Gumbel Softmax & 75.72\(^{\pm\text{.41}}\) \\
                 Hard Gumbel Softmax (\(\tau=1.0\times10^{-1}\)) & 75.85\(^{\pm\text{.31}}\) \\
                 Hard Gumbel Softmax (\(\tau=1.0\times10^{-3}\)) & 75.87\(^{\pm\text{.33}}\) \\
                 Hard Gumbel Softmax (\(\tau=1.0\times10^{-5}\)) & \textbf{76.33}\(^{\pm\textbf{.31}}\) \\
                 Hard Gumbel Softmax (\(\tau=1.0\times10^{-7}\)) & 76.02\(^{\pm\text{.50}}\) \\
            \end{tabular}
            }
        \end{minipage}
        \begin{minipage}{0.49\textwidth}
            \caption{Comparison of \(\ell_\mathrm{u}\) for MP-SSL}
            \label{tb:abl_unsupervised_loss}
            \resizebox{\textwidth}{!}{
            \begin{tabular}{lc}\toprule
                 \(\ell_\mathrm{u}\)& Cars Test Acc.(\%)\\\midrule
                 FreeMatch & 73.32\(^{\pm\text{.40}}\) \\
                 L1 Distance & 73.80\(^{\pm\text{.73}}\) \\
                 L2 Distance & 74.71\(^{\pm\text{.86}}\) \\
                 Smooth L1 Distance & 74.67\(^{\pm\text{.60}}\) \\
                 SCR (Eq.~(\ref{eq:scr})) & \textbf{76.33}\(^{\pm\textbf{.31}}\) 
            \end{tabular}
            }
        \end{minipage}
    \vspace{-3mm}
\end{table}

\subsubsection{Synthetic Consistency Regularization}\label{sec:eval_abl_SCR}
We lastly provide an ablation study of SCR defined by a cosine distance form as Eq.~(\ref{eq:scr}).
We tested four variants of \(\ell_\mathrm{u}\) in MP-SSL: (a) FreeMatch~\citep{Wang_ICLR23_freematch} that updates the entire model \(f_\theta\) including the classifier head \(h_\omega\), (b) L1 distance, i.e., \(|g_\psi(T_\mathrm{W}(\hat{x}_\mathrm{u})) - g_\psi(T_\mathrm{S}(\hat{x}_\mathrm{u}))|\), (c) L2 distance, i.e., \(\|g_\psi(T_\mathrm{W}(\hat{x}_\mathrm{u})) - g_\psi(T_\mathrm{S}(\hat{x}_\mathrm{u}))\|^2_2\), (d) Smooth L1 distance~\citep{Girshick_ICCV15_fastrcnn}.
We list the results in Table~\ref{tb:abl_unsupervised_loss}.
First, we see that our SCR loss significantly outperforms the FreeMatch loss.
This means that the consistency regularization on the feature spaces is quite effective for gSSL, as we expected in Sec.~\ref{sec:scr}.
Second, among the variants of SCR, the cosine distance based loss function achieved the best results.
We conjecture that losses that directly minimize differences between feature vectors, such as L1 and L2 distance, involve the L1 and L2 norm of the feature vector. Therefore, the norm of the feature vectors during training is relatively smaller, which hurts the norm of the loss gradients of classification tasks~\citep{Hariharan_ICCV17_low}.

\section{Related Work}
\paragraph{Semi-supervised Learning.}
Semi-supervised Learning (SSL) is a paradigm that trains a supervised model with labeled and unlabeled samples by simultaneously minimizing supervised and unsupervised loss.
Historically, various SSL algorithms have been proposed for deep learning such as entropy minimization~\citep{grandvalet_NIPS04_entmin}, pseudo-label~\citep{lee_ICMLW13_pseudo_labels}, and consistency regularization~\citep{bachman_NIPS14_learning_with_pseudo_ensembles,sajjadi_NIPS16_regularization_SSL,laine_ICLR16_temporal_ensembling}.
UDA~\citep{xie_NIPS20_UDA} and FixMatch~\citep{sohn_NIPS20_fixmatch}, which combine ideas of pseudo-label and consistency regularization, have achieved remarkable performance.
More recent methods such as FreeMatch~\citep{Wang_ICLR23_freematch} improve UDA and FixMatch to adaptively control the confidence threshold of acceptance of the pseudo labels for preventing error accumulation and overfitting.
These SSL algorithms assume that many unlabeled data are provided because unlabeled samples can be more easily obtained than labeled samples with human annotations.
However, we point out that even unlabelled data is becoming more difficult in today's increasingly privacy-conscious world.
This paper opens up a new SSL paradigm that makes unlabelled data unnecessary by leveraging pre-trained generative foundation models.\looseness-1

\paragraph{Leveraging Generative Models for Training Discriminative Models.}
In the context of data augmentation and transfer learning, several studies have applied the expressive power of conditional generative models to boost the performance of discriminative models, e.g., classifiers.
\cite{Zhu2018bmvc_CGAN_Augmentation}, \cite{yamaguchi_AAAI20_effective_data_augmentation_with_GANs}, \cite{Yamaguchi_NeurIPS23_MGR}, and \cite{He_ICLR23_synthetic} have exploited the generated images from conditional GANs and diffusion models for data augmentation and representation learning, and \cite{sankaranarayanan_CVPR18_domain_adaptation_gan} have introduced conditional GANs for domain adaptation setting to learn feature spaces of source and target domains jointly.
\cite{Li_CVPR20_unsupervised_domain_adaptation_without_source_data} have implemented an unsupervised domain adaptation technique with conditional GANs in a setting of no accessing source datasets.
More similar to our work, \cite{Yamaguchi_arXiv22_PSSL} have proposed a transfer learning method called P-SSL using pre-trained generative foundation models in semi-supervised learning.
However, we note that P-SSL and our method differ three-fold: (a) problem setting, (b) assumptions of data and label spaces, and (c) optimization methods.
For (a), the problem setting of our method is focused on SSL, whereas P-SSL is for transfer learning, where the neural architectures of source and target classifiers are different.
For (b), our method assumes the generative foundation model \(G_\mathrm{F}\) covers the training data space \(\mathcal{X}\). In contrast, P-SSL assumes the label space of \(G_\mathrm{F}\) covers the training label space i.e., \(\mathcal{Y}\subset\mathcal{Y}_\mathrm{F}\); the latter is more strict and thus the performance might degrade when it does not hold~\citep{Yamaguchi_arXiv22_PSSL}.
For (c), we directly optimize the latent variables of \(G_\mathrm{F}\) to find optimal unlabeled samples for SSL, whereas P-SSL just samples related synthetic samples via similarity in the label spaces through source pre-trained classifier.
These differences produce the performance improvements of our method in SSL, as shown in Sec.~\ref{sec:experiment}.

\section{Conclusion}
This paper presents a new semi-supervised learning (SSL) problem setting called generative SSL, where real unlabeled datasets are unavailable, where a generative foundation model is given as the source of unlabeled data.
This setting is important because we are often restricted from obtaining real unlabeled data due to privacy concerns.
To solve this problem, we propose a training method called MP-SSL, which consists of latent meta-optimization (LMO) and synthetic consistency regularization (SCR).
We experimentally demonstrate that our MP-SSL outperforms existing baselines and can potentially replace real unlabeled datasets with generative foundation models.
One of the limitations of this work is the dependency on the existence of foundation generative models, but this limitation will be relaxed because the foundation model trend is rapidly developing for various modalities in the community.
Important future steps are to speed up or avoid the computation of meta-learning in LMO and to extend our method to diffusion models, which produce synthetic samples with higher fidelity but require higher computational costs for sampling than GANs.


\bibliography{ref}

\appendix

\section{Extended Related Work}
\cite{dai2017good} and \cite{dhar2021universum} have shown semi-supervised learning methods based on generative models. A major difference between these and our work is the requirement to train a generative model on target datasets. It is well known that training generative models can be costly, unstable, and low-quality. In this sense, the existing methods have tackled the problems by improving the training generative models on target datasets. In contrast, our method skips this training by using generative foundation models and produces useful synthetic samples by latent meta optimization.

\section{Application to Medical Imaging}
To demonstrate the applicability, we further evaluated our method MP-SSL on the Chaoyang dataset, which is for a medical imaging task classifying cancers. Table~\ref{tb:chaoyang-food} shows that our method performs well in medical imaging.

\begin{table}[h]
    \centering
        \caption{
            Performance comparison of ResNet-18 classifiers.
            }
        \label{tb:chaoyang-food}
            \begin{tabular}{lccccc}\toprule
                Method / Dataset & Chaoyang & Food-101 \\
              \midrule
                Base Model    &  81.88\(^{\pm\text{.11}}\) & 77.43\(^{\pm\text{.03}}\) \\ \midrule
                \textbf{Oracle SSL (\(\mathcal{D} + \mathcal{D}_\mathrm{u}\))}\\
                FreeMatch &  81.24\(^{\pm\text{.29}}\) & 77.12\(^{\pm\text{.81}}\) \\\midrule
                \textbf{Generative SSL (\(\mathcal{D} + G_\mathrm{F}\))}\\
                Na\"ive gSSL (FreeMatch) &  81.60\(^{\pm\text{.76}}\) & 77.59\(^{\pm\text{.20}}\) \\
                P-SSL  &  81.28\(^{\pm\text{.76}}\) &  77.59\(^{\pm\text{.20}}\) \\
                MP-SSL (Ours) &  \textbf{82.53}\(^{\pm\textbf{.32}}\) & \textbf{78.59}\(^{\pm\textbf{.17}}\) \\
                \bottomrule
            \end{tabular}
 \end{table}

\section{Scalability on larger datasets and models.} In the real world, it is difficult to construct datasets with more than millions of samples, and thus, target datasets are basically small. Our experiments concentrated on the evaluation in such a realistic setting. Nevertheless, the meta-optimization of our method is done for each batch, so it works regardless of the dataset size; Table~\ref{tb:chaoyang-food} shows that our method is effective on a larger Food-101 ($\approx$ 100,000 samples). Also, Table~\ref{tb:arch} shows the scalability of our method for the larger architectures. However, since the proposed method requires backpropagation from the classifier during meta-training, the computational cost increases as the classifier size increases.

\section{Additional Results on 10\% Datasets}
In the main paper, we only showed the results on smaller datasets, where the proposed method is more effective.  
Here, we show the full evaluation results on 10\% datasets in Table~\ref{tb:10per}.
We see that our method outperforms the baselines for all datasets.

\section{Comparison to using real ImageNet}
Table~\ref{tb:trans-ssl} shows that our method outperforms the SSL with real ImageNet (Transfer SSL), indicating the synthetic samples from our meta-learning-based method are superior to real samples. This also justifies the use of a generative model for SSL.

\section{Ablation Study of $\mathcal{L}_\mathrm{gap}$}
As the alternative metric, we also tried maximum mean discrepancy (MMD), which is a measure of the distribution gap, but it does not improve mean squared error (MSE). Since MSE is the simplest to implement, we chose the form of Eq. (8) as our proposed method.

\begin{table}[t]
    \centering
        \caption{
            Top-1 Acc. (\%) of ResNet-18.}
        \label{tb:10per}
        \resizebox{\columnwidth}{!}{
            \begin{tabular}{lccccccccc}\toprule
                Method / Dataset & 10\%-Aircraft & 10\%-Birds & 10\%-Cars & 10\%-DTD & 10\%-Flower & 10\%-Pets  \\
              \midrule
                Base Model    &  12.63\(^{\pm\text{.61}}\) & 26.22\(^{\pm\text{.65}}\) &  19.74\(^{\pm\text{.15}}\) & 47.93\(^{\pm\text{.19}}\) & 50.44\(^{\pm\text{.57}}\) & 74.79\(^{\pm\text{.56}}\) \\ \midrule
                \textbf{Oracle SSL (\(\mathcal{D} + \mathcal{D}_\mathrm{u}\))}\\
                FreeMatch &  12.10\(^{\pm\text{.34}}\) & 25.70\(^{\pm\text{.47}}\) & 18.07\(^{\pm\text{.83}}\) & 46.08\(^{\pm\text{.52}}\) & 49.72\(^{\pm\text{.69}}\) & 75.35\(^{\pm\text{1.3}}\) \\\midrule
                \textbf{Generative SSL (\(\mathcal{D} + G_\mathrm{F}\))}\\
                P-SSL  &  12.98\(^{\pm\text{.28}}\) &  26.99\(^{\pm\text{.28}}\) & 21.78\(^{\pm\text{.31}}\) & 45.51\(^{\pm\text{.28}}\) & 50.07\(^{\pm\text{.11}}\) & 75.75\(^{\pm\text{1.2}}\)\\
                MP-SSL (Ours) &  \textbf{15.48}\(^{\pm\textbf{.33}}\) & \textbf{27.66}\(^{\pm\textbf{.13}}\) &  \textbf{24.62}\(^{\pm\textbf{.21}}\) & \textbf{49.27}\(^{\pm\textbf{.58}}\) & \textbf{54.78}\(^{\pm\textbf{.65}}\) & \textbf{76.65}\(^{\pm\textbf{.50}}\) \\
                \bottomrule
            \end{tabular}
        }
 \end{table}
 
\begin{table}[t]
    \centering
        \begin{minipage}{0.49\textwidth}
    \centering
        \caption{
            Top-1 Accuracy (\%) of ResNet-18.
            }
        \label{tb:trans-ssl}
            \begin{tabular}{lcccc}\toprule
                Method / Dataset & Cars  \\
              \midrule
                Base Model    &  71.62\(^{\pm\text{.30}}\) \\ \midrule
                \textbf{Oracle SSL (\(\mathcal{D} + \mathcal{D}_\mathrm{u}\))}\\
                FreeMatch &  \underline{82.73}\(^{\pm\text{.41}}\) \\
                SCR &  75.11\(^{\pm\text{.14}}\) \\\midrule
                \textbf{Transfer SSL (\(\mathcal{D} + \mathcal{D}_\mathrm{s}\))}\\
                FreeMatch &  72.76\(^{\pm\text{2.4}}\) \\
                SCR &  73.53\(^{\pm\text{.21}}\) \\\midrule
                \textbf{Generative SSL (\(\mathcal{D} + G_\mathrm{F}\))}\\
                Na\"ive gSSL (FreeMatch) &  73.67\(^{\pm\text{.67}}\) \\
                P-SSL  &  72.45\(^{\pm\text{.30}}\) \\
                MP-SSL (Ours) &  \textbf{76.33}\(^{\pm\textbf{.31}}\) \\
                \bottomrule
            \end{tabular}
        \end{minipage}
        \begin{minipage}{0.49\textwidth}
    \centering
        \caption{
            Top-1 Accuracy (\%) on Cars.
            }
        \label{tb:arch}
        \resizebox{\columnwidth}{!}{
            \begin{tabular}{lcccc}\toprule
                Method / Architecture & ResNet-50 & ResNet-101 \\
              \midrule
                Base Model    &  77.83\(^{\pm\text{.30}}\) & 78.24\(^{\pm\text{.28}}\)  \\ \midrule
                \textbf{Oracle SSL (\(\mathcal{D} + \mathcal{D}_\mathrm{u}\))}\\
                FreeMatch &  84.87\(^{\pm\text{.21}}\) & 85.68\(^{\pm\text{.09}}\) \\
                \textbf{Generative SSL (\(\mathcal{D} + G_\mathrm{F}\))}\\
                Na\"ive gSSL (FreeMatch) &  79.65\(^{\pm\text{.23}}\) & 80.48\(^{\pm\text{.37}}\) \\
                P-SSL  &  78.68\(^{\pm\text{.16}}\) & 78.10\(^{\pm\text{.20}}\) \\
                MP-SSL (Ours) &  \textbf{81.59}\(^{\pm\textbf{.55}}\) & \textbf{83.65}\(^{\pm\textbf{.26}}\) \\
                \bottomrule
            \end{tabular}
        }
        \vspace{+1mm}
        \caption{Comparison of $\mathcal{L}_\text{gap}$ for MP-SSL}
        \label{tb:abl_unsupervised_loss}
        \resizebox{0.6\textwidth}{!}{
        \begin{tabular}{lc}\toprule
             Loss form & Cars Test Acc.(\%)\\\midrule
             MSE (Eq.~(8)) & 76.33\(^{\pm\text{.31}}\) \\
             MMD & 76.12\(^{\pm\text{.77}}\) \\
        \end{tabular}
        }
        \end{minipage}
 \end{table}





\end{document}